\documentclass[11pt]{article}

\usepackage[preprint]{acl}
\usepackage{booktabs}

\usepackage{times}
\usepackage{latexsym}

\usepackage[T1]{fontenc}
\usepackage[utf8]{inputenc}

\usepackage{microtype}

\usepackage{inconsolata}

\usepackage{graphicx}
\usepackage{amssymb}

\usepackage{multirow}
\usepackage{multicol}
\usepackage{algorithm2e}
\usepackage{float}
\usepackage{placeins}
\usepackage{amsmath}

\title{Beyond Text: LLM-Based Dimensional Emotion Evaluation in Multimodal Dialogue}

\author{
  Yutong Hu \and Jinho Choi \\
  Emory University \\
  Atlanta, GA \\
  \texttt{yutong.hu@emory.edu, jinho.choi@emory.edu}
}

\begin{document}
\raggedbottom
\maketitle
\begin{abstract}
Emotion recognition in conversation has been widely studied, but applying Large Language Models (LLMs) to continuous dimensional emotion evaluation in multimodal dialogue remains largely unexplored. We propose an LLM-based framework that performs discrete emotion recognition and Valence–Arousal–Dominance (VAD) dimensional evaluation on IEMOCAP, incorporating acoustic cues as natural language descriptions following the SpeechCueLLM approach. We evaluate six models spanning the LLaMA, GPT, and Qwen families under zero-shot prompting, few-shot prompting, and LoRA fine-tuning. LoRA fine-tuned LLaMA models substantially outperform prompt-engineered GPT models on both tasks despite GPT's larger scale, a gap we attribute to domain adaptation rather than model capacity. Our best model achieves a Valence CCC of 0.7822, a new state-of-the-art on IEMOCAP. Ablation studies confirm that textual audio descriptions meaningfully improve smaller models (+3.5–3.6 weighted F1) while contributing little for the largest model, suggesting audio cues are most valuable when linguistic capacity is limited. The performance asymmetry across VAD dimensions closely mirrors the annotator agreement hierarchy in IEMOCAP's own annotations.
\end{abstract}

\section{Introduction} \label{Introduction}

Emotion recognition in conversation (ERC) has gained attention for the past two decades. Unlike isolated utterance-level analysis, ERC requires the contextual understanding of utterances across multiple conversational turns with dynamic emotions \cite{wu_multimodal_2025}. This conversational dependency makes the problem substantially more challenging and more consequential for real-world applications such as mental health support and human–computer interaction.

There are two common ways for emotion representation. The first way is a discrete emotion label, which assigns a single categorical label(e.g. happiness, frustration, etc.) to an utterance. The second one is continuous dimensional emotions, which quantify emotional experience along continuous scales. And among the continuous dimensions, the Valence-Arousal-Dominance (VAD) system \cite{russell_evidence_1977} is the most widely adopted, measuring how positive, activated, and dominant a person feels. While discrete ERC has received the bulk of the attention \cite{lei_instructerc_2023, wu_beyond_2025}, dimensional evaluation has grown more recently, driven by the demand for richer emotional characterization.

Because emotions are not conveyed only by words alone, the incorporation of multimodal information is critical for both tasks: tone, pitch, rhythm, and speaking rate all carry affective signals that text transcriptions discard. The involvement of multimodality is therefore essential not only for the accuracy but also for the diverse potential downstream applications, such as emotion-conditioned speech generation. 

Currently, for multimodal ERC on VAD systems, Large Language Models (LLM) are barely used in this regime; instead, most dimensional work on multimodal dataset still relies on Deep Learning(DL) models, including LSTM and CNN-1D \cite{atmaja_two-stage_2021,messaoudi_speech_2024}, or pre-trained models such as HuBERT or DeBERTa, limiting the contextual reasoning capacity to exploit \cite{srinivasan_representation_2022,ispas_multi-task_2023}. To address this gap, we make the following contributions:
\begin{itemize}
\item We propose the first systematic LLM-based framework for continuous VAD dimensional emotion evaluation in multimodal dialogue, extending the SpeechCueLLM approach of incorporating acoustic cues as natural language descriptions from discrete ER to the dimensional setting.
\item We conduct a comprehensive comparison across six models spanning the LLaMA, GPT, and Qwen families under prompt engineering and LoRA fine-tuning, showing that parameter-efficient fine-tuning substantially outperforms prompting-based approaches for both discrete and dimensional emotion tasks, despite the larger scale of the prompted models.
\item Through error analysis and ablation studies, we identify the sources of the performance gap between fine-tuned and prompted models, the contribution of acoustic descriptions to performance, and the effect of past VAD context on prediction quality.

\end{itemize}
\section{Background and Related Work} \label{Background and Related Work}

\subsection{ Discrete Emotion Recognition in Dialogue }\label{Discrete Emotion Recognition in Dialogue}
Early DL approaches to Emotion Recognition (ER) employed CNNs for end-to-end speech representations~ \cite{trigeorgis_adieu_2016}, LSTMs for sequential dialogue modeling~ \cite{poria_context-dependent_2017}, and graph-based architectures~ \cite{ghosal_dialoguegcn_2019}. DialogueRNN~ \cite{majumder_dialoguernn_2019}, which models speaker state, emotion state, and global context via an attentive RNN, remains one of the most influential baselines for ERC on IEMOCAP and MELD. More recently, InstructERC~ \cite{lei_instructerc_2023} reformulates ERC as an instruction-following task using multi-task retrieval-augmented prompting. SpeechCueLLM~ \cite{wu_beyond_2025} extends this LLM-based approach to the multimodal setting by incorporating vocal information as natural language descriptions of acoustic features. This work directly adopted that framework for acoustic information incorporation.

\subsection{Dimensional Emotion Evaluation}\label{Dimensional Emotion Evaluation}
The Valence–Arousal–Dominance (VAD) framework~ \cite{russell_evidence_1977} provides a richer characterization of emotional state than discrete labels, capturing the positive–negative, activation–deactivation, and power axes respectively. Multimodal approaches combining audio and text for VAD prediction on IEMOCAP, using architectures such as LSTM/CNN-1D~ \cite{atmaja_two-stage_2021,awatef_multimodal_2025}, have advanced performance substantially. However, LLMs remain largely absent from this regime; the contextual reasoning capacity that drives gains in discrete ERC has not been systematically evaluated for continuous VAD prediction in multimodal dialogue, which is the gap this work addresses.

\subsection{Available Multimodal Models for Audio and Text Data}\label{Available Multimodal Models for Audio and Text Data}

Models capable of jointly processing audio and text fall into two broad categories: audio-language foundation models (e.g., Wav2Vec 2.0~ \cite{baevski_wav2vec_2020}, CLAP~ \cite{elizalde_clap_2022}, and SpeechT5~ \cite{ao_speecht5_2022}), which learn general-purpose audio representations; end-to-end conversational multimodal LLMs (e.g., Qwen3-Omni, GPT-5.1, and Gemini 3 Pro), which map speech directly into semantic representations within unified dialogue frameworks. This work does not adopt the latter due to their substantial computational requirements and uncertain impact on specialized ER tasks, leaving their exploration to future work.

\section{Methods}\label{Methods}

\subsection{Framework Overview}\label{Framework Overview}
\begin{figure*}
    \centering
    \includegraphics[width =0.94\linewidth]{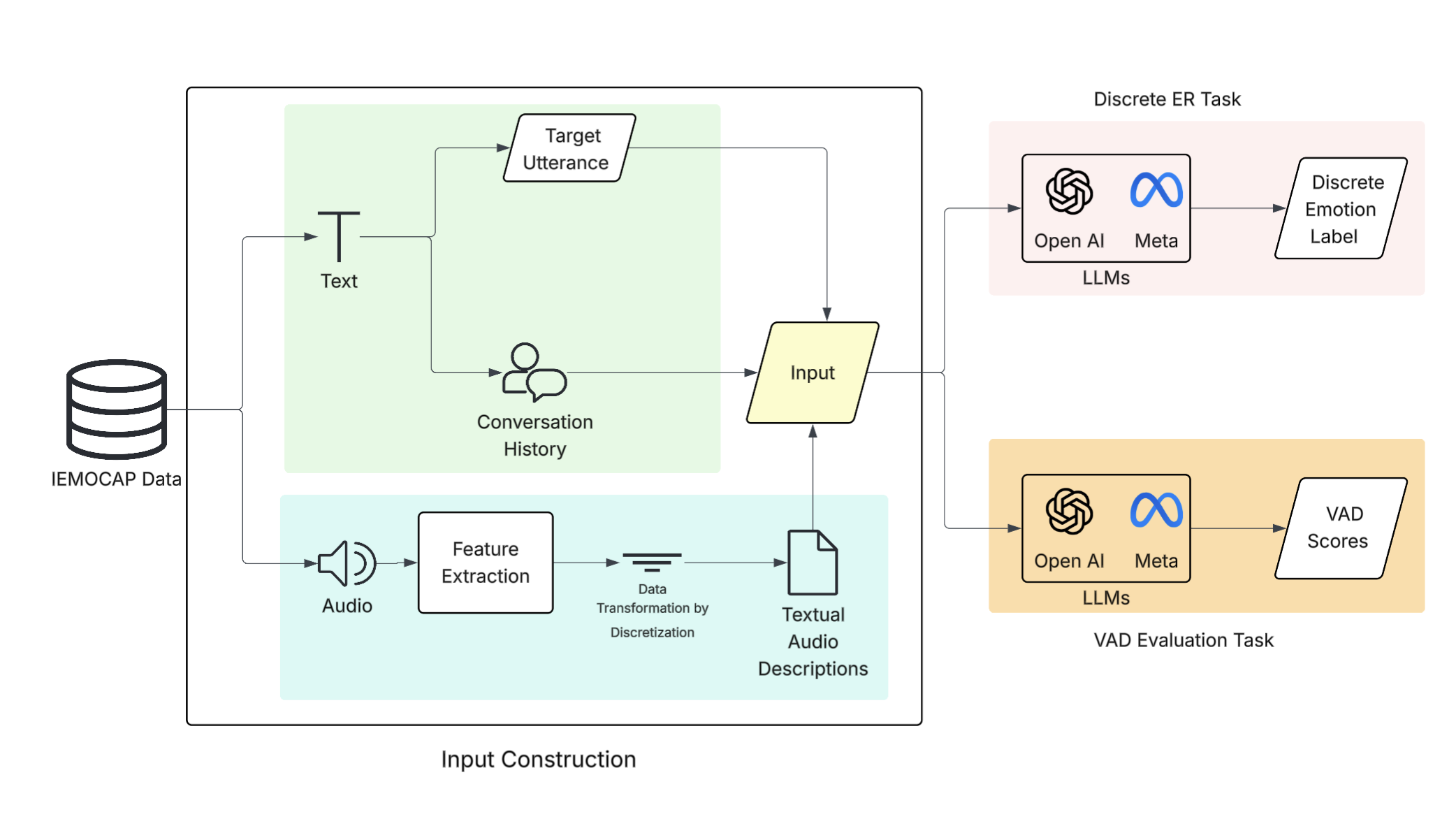}
    \caption{Overall Pipeline}
    \label{fig:pipeline}
\end{figure*}
This work proposes an LLM-based framework for performing discrete ER and continuous VAD dimensional evaluation on utterances in multimodal dialogues. While those two tasks share the same input construction strategy, they are treated as independent tasks, each with its own prompt formulation, and in the case of the LoRA fine-tuning, its own separately trained model instance adapted to the corresponding annotation type. The framework takes three sources of information as input: the conversation history, the target utterance, and the audio feature descriptions. A discrete emotion label will be output for the discrete ER, while a set of VAD scores (the representation the emotional state along Valence, Arousal, and Dominance dimensions) will be output for the continuous emotion evaluation.

Figure \ref{fig:pipeline} and the algorithm \ref{alg:framework} illustrate the overall pipeline. Given a target utterance from IEMOCAP, the framework retrieves the preceding structured conversation (up to 12 utterances in this study) to provide dialogue context. Simultaneously, acoustic features are extracted from the audio recording of the target utterance and converted to natural language descriptions following SpeechCueLLM approach \cite{wu_beyond_2025}. Based on the required task and passed structured text prompts with all required inputs, the models will output either the discrete emotion category selected from a predefined label set (happy, sad, neutral, angry, excited, and frustrated) or numerical integer VAD scores on a 1-5 scale. The framework is designed to allow direct comparison across models of different scales and training paradigms. In this work, we evaluate the framework mainly under two strategies: prompt engineering (zero-shot and few-shot) and LoRA fine-tuning. 
\subsection{Audio Feature Extraction and Textual Description}\label{Audio Feature Extraction and Textual Description}

A central design choice of this framework is to incorporate acoustic information in natural language descriptions rather than raw audio features or learned audio embeddings. This approach, adopted from SpeechCueLLM \cite{wu_beyond_2025}, enables LLMs to access audio information that is included in the input without requiring architectural modification to handle the multimodal input directly.

For each utterance in this dataset, three acoustic features are extracted from the raw audio recording: the perceived volume, pitch, and speaking rate. For each of the features, both the central tendency and the degree of variation within the utterances are computed, yielding six descriptive values in total for each utterance.

To convert these acoustic measurements into natural language descriptions, a quantile-based scheme is utilized to assign one of the categorical labels to acoustic feature values and variation measurements: very low, low, moderate, high, or very high. This produces a concise and interpretable description of each utterance's acoustic information as in Figure \ref{fig:audio_feature_desc_eg}.

This textual audio description provide models with non-lexical cues that are not recoverable from the transcribed text alone. And the contribution of the audio description to model performance is evaluated through an ablation study comparing models with or without them, the results of which are discussed in Section \ref{The Impact of Acoustic Feature Descriptions}.

\subsection{Fine-Tuning Strategy (LoRA)}\label{Fine-Tuning Strategy (LoRA)}
While LLMs possess strong general language understanding capabilities, adaption to the target domain and annotation conventions is often required for their effective application to specialized tasks. Full fine-tuning of LLMs is computationally expensive given the scale of the modern models, as it requires updating all model parameters simultaneously. To address this, we adopt Low-Rank Adaptation(LoRA), a parameter-efficient fine-tuning method that introduces a small number of trainable parameters while keeping the original model weights frozen \cite{hu_lora_2021}.

The IEMOCAP dataset, while the most appropriate available resource for this task, provides a limited number of training samples compared to the scale of the datasets those LLMs were originally trained on. LoRA's constrained parameterization reduces the risk of overfitting under the data-limited situation.

\section{Experiment Setup}\label{Experiment Setup}

\subsection{Dataset}\label{Dataset}

This study requires the dataset to satisfy: (I) multimodal, containing at least audio and text modalities; (II) conversational; (III) annotated with both discrete and continuous emotion. While textual emotion datasets with discrete labels are relatively common, datasets meeting all three criteria are rare. After reviewing 19 emotion datasets in total (see Table \ref{tab:emotion_datasets} in  Appendix \ref{Multimodal Emotion Dataset}), we chose IEMOCAP \cite{busso_iemocap_2008}, one of the most widely used datasets for multimodal emotion research.

\subsubsection{IEMOCAP Dataset}\label{IEMOCAP Dataset}

The Interactive Emotional Dyadic Motion Capture (IEMOCAP) \cite{busso_iemocap_2008} dataset is a multimodal corpus designed for the study of expressive human communication in dyadic interaction. The dataset consists of 151 dyadic dialogues performed by 10 actors arranged in five sessions, each pairing one male and one female actor. The recordings include both scripted scenarios and improvised interactions elicited through emotional prompts, yielding a total of 10,086 utterances with an average of approximately 66 utterances per dialogue.

The dataset captures four modalities for each utterance: audio recordings, video recordings, text transcriptions, and motion capture data tracking facial and hand movements. The total duration of all audio recordings is approximately 12 hours. In this work, we only utilized the audio recordings and text transcriptions, as these modalities are most directly relevant to the proposed framework and align with the SpeechCueLLM approach adopted for audio feature extraction.

Each utterance in IEMOCAP is annotated by multiple evaluators with a discrete 
emotion category. The final label is determined by majority voting across 
annotators. The original label set covers nine 
categories: anger, sadness, frustration, happiness, excitement, neutral state, 
surprise, fear, and other. In addition to categorical labels, annotators 
provided VAD ratings on a scale of 1 to 5, where higher values indicate more 
positive valence, higher arousal, and greater dominance respectively. The aggregated annotator VAD scores were used as the ground truth for each utterance in this work.

For the discrete ER task, we followed common practice in the literature \cite{wu_beyond_2025,lei_instructerc_2023}, excluding low-frequency categories Surprise, Fear, and Others from the label set and arriving at six discrete emotion categories and a total of 7,433 utterances.

\subsubsection{Data Splitting}\label{Data Splitting}
To evaluate model generalization across speakers, we adopted a Leave-One-Subject-Out (LOSO) splitting strategy. Specifically, the fifth session of IEMOCAP was held out as the test set, as it contains unseen speakers not present in the training data, ensuring the model is evaluated on its ability to generalize to new speakers. The remaining four sessions were used for training and validation, split at a 90/10 ratio respectively. This splitting strategy is consistent with prior work on IEMOCAP.

\subsection{Models}\label{Models}
We evaluated the proposed framework across six language models spanning three families (the Qwen model was only used for discrete ER evaluation), enabling comparison between open-source and closed-source systems as well as across different parameter scales.

Within the open-source models, we selected three models from the LLaMA series: LLaMA-2-7B, LLaMA-3.1-8B, and LLaMA-3.3-70B. LLaMA-2-7B and LLaMA-3.1-8B represent smaller, more computationally accessible configurations, while LLaMA-3.3-70B represents a large-scale model. To evaluate more comprehensively, we also included the Qwen3.5-35B-A3B model for the discrete ER task. All three LLaMA models were evaluated under zero-shot prompting, few-shot prompting, and LoRA fine-tuning conditions for discrete ER, and only LoRA fine-tuning for VAD evaluation. 

For closed-source comparison, we included two OpenAI models: GPT-4o-mini and GPT-5-mini. GPT models are evaluated under zero-shot and few-shot prompting conditions only. 

\subsection{Implementation Details}\label{Implementation Details}
For LoRA fine-tuning, we used the AdamW optimizer with a learning rate of 0.0003. The LoRA rank is set to $r$ = 16 with the scaling parameter $\alpha$ set equal to $r$, following standard practice. All models are trained for 15 epochs.

For the LLaMA-3.3-70B model, training was conducted using DeepSpeed to handle the memory requirement of the 70-billion-parameters configurations across two GPUs. Smaller models were trained on a single GPU configuration. 

For prompt engineering conditions, zero-shot prompts follow the templates described in Appendix \ref{Prompt Templates}  Few-shot prompts prepend a fixed set of annotated examples to the same template.

\section{Evaluation Metrics}\label{Evaluation Metrics}

\subsection{Discrete ER}\label{Discrete ER}
The discrete emotion recognition task was evaluated using the weighted F1 score, defined as:

\begin{equation}
    F1 = \frac{2 \times \text{Precision} \times \text{Recall}}{\text{Precision} + \text{Recall}}
\end{equation}

\begin{equation}
    \text{Weighted F1} = \sum_{i =1}^{N} w_i \times \text{F1}_i
\end{equation}

\noindent where $w_i$ is the proportion of samples belonging to class $i$ and $\text{F1}_i$ is the F1 score for that class. The weighted F1 score is chosen over macro F1 because it accounts for class imbalance in the IEMOCAP dataset, where emotion categories are not uniformly distributed.

\subsection{VAD Evaluation}\label{VAD Evaluation}
The VAD evaluation task was assessed using the Concordance Correlation Coefficient (CCC), which jointly captures both the correlation and the 
agreement between predicted and ground truth values by penalizing mean offset in 
addition to variance differences:

\begin{equation}
    \text{CCC} = \frac{2r\sigma_x\sigma_y}
    {\sigma_x^2 + \sigma_y^2 + (\mu_x - \mu_y)^2}
\end{equation}

\noindent where $r$ is the Pearson correlation, $\sigma_x$ and $\sigma_y$ are the standard deviations, and $\mu_x$ and $\mu_y$ are the means of the predicted and ground truth distributions respectively. CCC ranges from $-1$ to $1$, where $1$ indicates perfect agreement.

\section{Emotion Recognition Results}\label{Emotion Recognition Results}
Table \ref{tab:result_der} reports weighted F1 scores for discrete ER across all six models and training/prompting conditions, including comparison with SpeechCueLLM \cite{wu_beyond_2025} baseline. 
\FloatBarrier
\begin{table*}[t!]
    \centering
    \resizebox{\linewidth}{!}{\begin{tabular}{|c|l|l|l|l|l|l|l|}\hline

 & LLaMA-2-7B&   LLaMA-3.1-8B& LLaMA-3.3-70B& Qwen3.5-35B-A3B&GPT 4o mini & GPT 5 mini&SpeechCueLLM \cite{wu_beyond_2025}\\\hline
         Zero-shot&   9.058&   31.293& \textbf{60.299}& 14.317& 54.700 & 58.490&-\\\hline
         Few-shot& 
     25.675&   38.762& 58.280& 21.832& 56.822 & \textbf{59.600}&-\\ \hline
 LoRA& \textbf{73.196}& 71.818& 72.122 & 68.493&-&-&72.021 \\\hline
 
 \end{tabular}}
    \caption{Discrete ER Performance of All Models (Weighted F1 score)}
    \label{tab:result_der}
\end{table*}

The most prominent finding is the LoRA fine-tuned open-source models consistently outperform the prompt-engineered ones. With all three LoRA fine-tuned LLaMA models achieving scores between 71.8 and 73.2 along with the fine-tuned Qwen3.5 model of 68.493, there is a large gap between prompt engineering and LoRA fine-tuning for open-source models used here. Among prompt engineering conditions, LLaMA-3.3-70B zero-shot (60.299) performs comparably to GPT-4o-mini few-shot (56.8), suggesting that scale partially compensates for the absence of task-specific adaptation. Fine-tuned models uniformly surpass all prompting-based approaches, including GPT-5-mini few-shot (59.6).

\section{VAD Evaluation Results}\label{VAD Evaluation Results}
Table \ref{tab:result_vad} reports VAD evaluation performance for prompt-engineered GPT models and LoRA fine-tuned LLaMA models. Due to the huge gap between the performance of LLaMA models with LoRA fine-tuning and with the prompt engineering, we only applied LoRA fine-tuning for LLaMA models within the experiment for VAD evaluation.

\begin{table*}[ht]
\centering
\caption{VAD Evaluation Performance Comparison Across Different Models}
\label{tab:result_vad}
\scalebox{0.85}{
\begin{tabular}{ll|cc|cc|ccc}
\toprule
& & \multicolumn{2}{c}{GPT-4o mini} & \multicolumn{2}{c}{GPT-5 mini} &\multicolumn{3}{c}{LoRA Fine-tuned LLaMA Model}\\
\cmidrule(lr){3-4} \cmidrule(lr){5-6}\cmidrule(lr){7-9}
&Metric & Zero-shot & Few-shot & Zero-shot & Few-shot &2-7B&3.1-8B&3.370B \\
\midrule

& Valence       & 0.6024 & 0.6311 & 0.6630 & 0.6697 &0.7672&0.7433 & \textbf{0.7822}\\

& Arousal       & 0.3393 & 0.3589 & 0.3926 & 0.3416 &0.4406&\textbf{0.4778} &0.4653 \\

& Dominance       & 0.1458 & 0.1235 & 0.2990 & 0.3046&0.4388&0.4400& \textbf{0.4413} \\

& Overall  & 0.3625 & 0.3712 & 0.4515 & 0.4386&0.5489&0.5537 &\textbf{0.5629} \\
\bottomrule
\end{tabular}}
\end{table*}
Among GPT models, GPT-5-mini outperforms GPT-4o-mini across most conditions. The most pronounced difference appears on the Dominance dimension, where GPT-5-mini zero-shot achieves a CCC of 0.299 versus 0.146 for GPT-4o-mini zero-shot. Valence is the best-predicted dimension across all prompt engineering conditions (CCC 0.60–0.67), while Arousal and Dominance remain substantially lower (CCC 0.13–0.39).

LoRA fine-tuned LLaMA models substantially outperform all prompt-engineered approaches. On Valence, LLaMA-3.3-70B achieves a CCC of 0.7822, leading the best GPT-5-mini prompting result of 0.6697 by around 0.11. LLaMA-2-7B and LLaMA-3.1-8B also achieve strong Valence CCC values of 0.7672 and 0.7433 respectively, demonstrating that the gain is not primarily attributable to model scale. On Arousal and Dominance, LoRA fine-tuned models also outperform GPT prompting (CCC values from 0.44 to 0.48 vs. from 0.34 to 0.39 for Arousal and around 0.44 vs. from 0.12 to 0.30 for Dominance), though absolute values remain substantially lower than Valence.

\begin{table*}[]
    \centering
    \scalebox{0.9}{
    \begin{tabular}{|c|l|l|l|l|} \hline
    
       & Modalities & Valence CCC& Arousal CCC& Dominance CCC \\\hline
         Atmaja et al. \cite{atmaja_two-stage_2021}  & Audio, Text& 0.553&0.579&0.456\\\hline
         Messaoudi et al. \cite{messaoudi_speech_2024}&Audio&0.236&0.571&0.408\\\hline
         Awatef et al. \cite{awatef_multimodal_2025}&Audio, Text&0.603&\textbf{0.736}&\textbf{0.604}\\\hline
         The Proposed&Audio, Text&\textbf{0.782}&0.465&0.441\\\hline
         
    \end{tabular}}
    The proposed here is LLaMA-3.3-70B with LoRA fine-tuning
    \caption{Comparison with Prior Work on VAD Evaluation on IEMOCAP}
    \label{tab:comparison_vad}
\end{table*}

Table \ref{tab:comparison_vad}
compares our best model against prior works on VAD evaluation on IEMOCAP; our approach achieves state-of-the-art performance on the Valence dimension, surpassing all baselines by a substantial margin, while Arousal and Dominance remain competitive but below the best reported results.

Among those previous works, the one by Awatef et al. \cite{awatef_multimodal_2025}, even though it has a lower Valence CCC, outperforms our model on both Arousal (0.736 vs. 0.465) and Dominance (0.604 vs. 0.441) by a notable gap. Their work utilized LSTM and CNN-1D models with a late fusion by Neural Network. While their architecture was trained end-to-end on acoustic features, our approach used discretized textual audio descriptions. The Valence gain suggests LLMs bring complementary contextual reasoning that compensates on the dimension with highest annotator agreement.

\section{Error Analysis for Discrete ER}\label{Error Analysis for Discrete ER}
While the substantial performance gap between LoRA fine-tuned LLaMA models and prompt-engineered GPT models is somewhat unexpected given the latter's considerably larger scale, a closer examination of per-emotion F1 scores and confusion patterns helps explain this discrepancy. 
Tables \ref{tab:per_emotion} and \ref{tab:confusions}
 present the per-emotion F1 scores and key bidirectional confusion rates respectively.
 \begin{table}[ht]
\centering
\caption{Per-emotion F1 score (\%)}
\label{tab:per_emotion}
\resizebox{\linewidth}{!}{
\begin{tabular}{lcccccc}
\toprule
Model & Happy & Sad & Neutral & Angry & Excited & Frustrated \\
\midrule
GPT-4o-mini ZS & 49 & 70 & 53 & 32 & 53 & 60 \\
GPT-4o-mini FS & 46 & 74 & 55 & 40 & 56 & 60 \\
GPT-5-mini ZS  & 42 & 70 & 57 & 45 & 63 & 61 \\
GPT-5-mini FS  & 43 & 70 & 60 & 53 & 61 & 61 \\
LLaMA-2-7B ZS   & 28 & 2 & 4 & 3 & 5 & 8 \\
LLaMA-2-7B FS   & 40 & 3 & 5 & 4 & 9 & 7 \\
LLaMA-3.1-8B ZS & 41 & 55 & 52 & 22 & 6 & 15 \\
LLaMA-3.1-8B FS & 44 & 42 & 54 & 38 & 24 & 37 \\
LLaMA-3.3-70B ZS & 50 & 78 & 31 & 70 & 67 & 40 \\
LLaMA-3.3-70B FS & 42 & 74 & 58 & 44 & 57 & 56 \\
\midrule
FT LLaMA-2-7B   & 63 & 83 & 75 & 69 & 72 & 72 \\
FT LLaMA-3.1-8B & 66 & 84 & 74 & 61 & 71 & 69 \\
FT LLaMA-3.3-70B & 61 & 80 & 73 & 65 & 78 & 69 \\
\bottomrule
\end{tabular}}\\
ZS: zero-shot; FS: few-shot; FT: fine-tuned.
\end{table}

Based on Table \ref{tab:per_emotion}, all models (despite LLaMA-2-7B and LLaMA-3.1-8B with few-shot) have the best performance for Sad emotion. The prompt-engineered GPT models are outperformed by fine-tuned LLaMA models under every emotion category. For GPT models, Happy and Angry are the two discrete emotions where they have the lowest F1 scores. Few-shot prompting generally slightly improved GPT models' performance over zero-shot prompting. GPT-4o-mini gets similar scores to GPT-5-mini for Frustrated and Sad, slightly outperforms it in Happy, and underperforms for Neutral, Angry, and Excited.
\begin{table*}[ht]
\centering
\caption{Key Bidirectional Confusion Rates (\%)}
\label{tab:confusions}
\scalebox{0.7}{
\begin{tabular}{lcccccc}
\toprule
& \multicolumn{2}{c}{Happy $\leftrightarrow$ Excited} 
& \multicolumn{2}{c}{Angry $\leftrightarrow$ Frustrated}
& \multicolumn{2}{c}{Sad $\leftrightarrow$ Frustrated} \\
\cmidrule(lr){2-3}\cmidrule(lr){4-5}\cmidrule(lr){6-7}
Model & Hap$\to$Exc & Exc$\to$Hap & Ang$\to$Fru & Fru$\to$Ang & Sad$\to$Fru & Fru$\to$Sad \\
\midrule
GPT-4o-mini ZS & 7.6 & 29.8 & 75.3 & 3.1 & 22.0 & 3.4 \\
GPT-4o-mini FS & 12.5 & 22.7 & 65.9 & 6.3 & 15.1 & 6.0 \\
GPT-5-mini ZS  & 30.6 & 18.4 & 60.0 & 6.3 & 18.8 & 4.5 \\
GPT-5-mini FS  & 29.2 & 22.1 & 46.5 & 12.1 & 11.0 & 8.7 \\
\midrule
LLaMA-2-7B ZS   & 1.1 & 81.5 & 10.2 & 3.3 & 2.5 & 10.4 \\
LLaMA-2-7B FS   & 5.1& 86.7 & 18.4 & 5.1 & 6.5 & 5.7 \\
LLaMA-3.1-8B ZS & 1.4 & 67.2 & 9.4 & 1.0 & 1.8 & 12.6 \\
LLaMA-3.1-8B FS & 4.9 & 52.8 & 22.4 & 5.2 & 6.1 & 6.8 \\
LLaMA-3.3-70B ZS & 34.2 & 17.8 & 16.7 & 17.6 & 4.2 & 36.5 \\
LLaMA-3.3-70B FS & 18.7 & 17.0 & 56.8 & 5.8 & 15.6 & 5.5 \\
\midrule
FT LLaMA-2-7B   & 21.5 & 17.7 & 27.6 & 12.1 & 13.1 & 1.8 \\
FT LLaMA-3.1-8B & 11.8 & 20.7 & 37.1 & 10.0 & 9.4  & 3.9 \\
FT LLaMA-3.3-70B & 34.0 & 10.7 & 30.0 & 14.7 & 12.2 & 2.1 \\
\bottomrule
\end{tabular}}

ZS: zero-shot; FS: few-shot; FT: fine-tuned.
\end{table*}
Table \ref{tab:confusions} implies the potential reason for GPT models not meeting our expectations. GPT models have high confusion rates on angry and frustrated. The confusion rates of 46.5-75.3 for GPT models on misinterpreting anger as frustration show they are less sensitive to aggressiveness. One interesting difference between GPT-4o-mini and GPT-5-mini models is that GPT-5-mini tends to overestimate the energy of positive emotions (confusion rate of 29.2-30.6 on Happy$\rightarrow$Excite vs. 18.4-22.1 for Excite$\rightarrow$Happy), while GPT-4o-mini, in contrast, is more conservative about the positive levels (confusion rate of 7.6-12.5 on Happy$\rightarrow$Excite vs. 22.7-29.8 for Excite$\rightarrow$Happy). That explains why LoRA fine-tuned LLaMA models outperform GPT models here.

The poor performance of LLaMA-2-7B and LLaMA-3.1-8B models with prompt engineering can also be explained now. They tend to collapse most emotions into a single label, as presented by the extremely high ratio of excitement instances misclassified as happiness and the low confusion rate over other emotion label pairs.
\section{VAD Evaluation Performance Analysis}\label{VAD Evaluation Performance Analysis}
A consistent hierarchy emerges across all model families and training conditions: Valence is predicted substantially better than Arousal, which is predicted at roughly the same level as Dominance. This pattern holds both for GPT prompt-engineered models (Valence CCC 0.60–0.67 vs. Arousal CCC 0.34–0.39) and for LoRA fine-tuned LLaMA models (Valence CCC 0.74–0.78 vs. Arousal CCC 0.44–0.48). The gap is particularly striking given that all three dimensions share the same scale, annotation protocols, and model architecture.

This performance hierarchy directly mirrors the annotator agreement hierarchy reported in Table \ref{tab:annotator}. Valence has the highest inter-annotator agreement (Krippendorff's $\alpha$ = 0.680), while Dominance has the lowest ($\alpha$ = 0.271) and Arousal is intermediate ($\alpha$ = 0.304). When annotators themselves disagree substantially, the annotation ground truth is inherently noisy, imposing a natural ceiling on achievable model performance. The agreement hierarchy--Valence, Arousal, Dominance-- is consistent with the VAD evaluation performance.

\section{Annotator Reliability}\label{Annotator Reliability}
Because emotions are subjective, annotators are likely to have different answers for the same target sentence. We conducted an inter-agreement measurement with Krippendorff's $\alpha$ and Fleiss' $\kappa$ to test the reliability of data. Table \ref{tab:annotator}
 reports the test results for each annotation type in IEMOCAP. 
\begin{table}[]
    \centering
    \resizebox{\linewidth}{!}{
    \begin{tabular}{|c|l|l|}\hline
 & \multicolumn{2}{|c|}{Reliability Measurements }\\\hline
  & Krippendorff's alpha&Fleiss' Kappa\\\hline
 
 Discrete Emotion Labels&0.279&0.273\\
 \hline
 Valence Score&0.680&0.316\\
 \hline
 Arousal Score&0.304&0.085\\
 \hline
 Dominance Score&0.271&0.014 \\
 \hline
        
\end{tabular}}
    \caption{Reliability Measurements of IEMOCAP}
    \label{tab:annotator}
\end{table}

 Discrete emotion labels achieve only fair agreement ($\alpha$ = 0.279 and $\kappa$ = 0.273), reflecting inherent ambiguity in emotion assignment. Valence shows substantially higher agreement ($\alpha$ = 0.680). Arousal shows intermediate agreement ($\alpha$ = 0.304 and $\kappa$ = 0.085), comparable to discrete labels. Dominance shows near-floor alpha and also near-zero Kappa ($\alpha$ = 0.271 and $\kappa$ = 0.014), suggesting that while annotators show some ordinal agreement on dominance, they diverge substantially on absolute scores. These reliability statistics contextualize the model performance results for VAD evaluation.

\section{Ablation Studies}\label{Ablation Studies}
\subsection{The Impact of Acoustic Feature Descriptions}\label{The Impact of Acoustic Feature Descriptions}
Natural language audio description is part of the input to LLMs. We also conducted experiments to reveal the contribution of audio feature description to final performance by removing audio feature descriptions in discrete ER performance for LoRA fine-tuned LLaMA models. Table \ref{tab:audio_description}
 reports this impact driven by the removal. 
 \begin{table*}[]
    \centering
    \scalebox{0.8}{
    \begin{tabular}{|c|l|l|l|}\hline

 & LLaMA-2-7B&   LLaMA-3.1-8B& LLaMA-3.3-70B\\\hline
         
 LoRA& \textbf{73.196}& 71.818& 72.122  \\\hline
 LoRA without audio description& 69.672 &68.23&72.108
 \\ \hline
 \end{tabular}}
    \caption{The Impact of Audio Description on Discrete ER Performance of LLaMA Models(Weighted F1 score)}
    \label{tab:audio_description}
\end{table*}

Removing audio descriptions reduces weighted F1 by 3.5 for LLaMA-2-7B (73.2 vs. 69.7) and 3.6 for LLaMA-3.1-8B (71.8 vs. 68.2), while having negligible effect on LLaMA-3.3-70B (72.1 vs. 72.1). This suggests that audio information provides useful signal for smaller models but that the larger model can partially recover prosodic context from linguistic patterns.

\subsection{The Impact of Past VAD Context}\label{The Impact of Past VAD Context}
Another interesting question here is whether LLMs provided with the VAD scores for utterances in conversation history will improve their performance. Here the VAD scores for previous utterances are the outputs predicted by models themselves rather than the ground truth VAD values.
\subsubsection{GPT Models with Prompting}\label{GPT Models with Prompting}
Table \ref{tab:gpt_past_vad12} reports VAD evaluation performance for GPT models when past VAD scores are provided in the context window (12 and 3 utterances respectively). Adding past VAD context yields modest improvements for GPT-5-mini on Valence (CCC of 0.663 to 0.689 with window =12) but has minimal or negative effects on Arousal and Dominance. The window size makes little difference overall.
\begin{figure*}[t]
    \centering
    \includegraphics[width =0.78\linewidth]{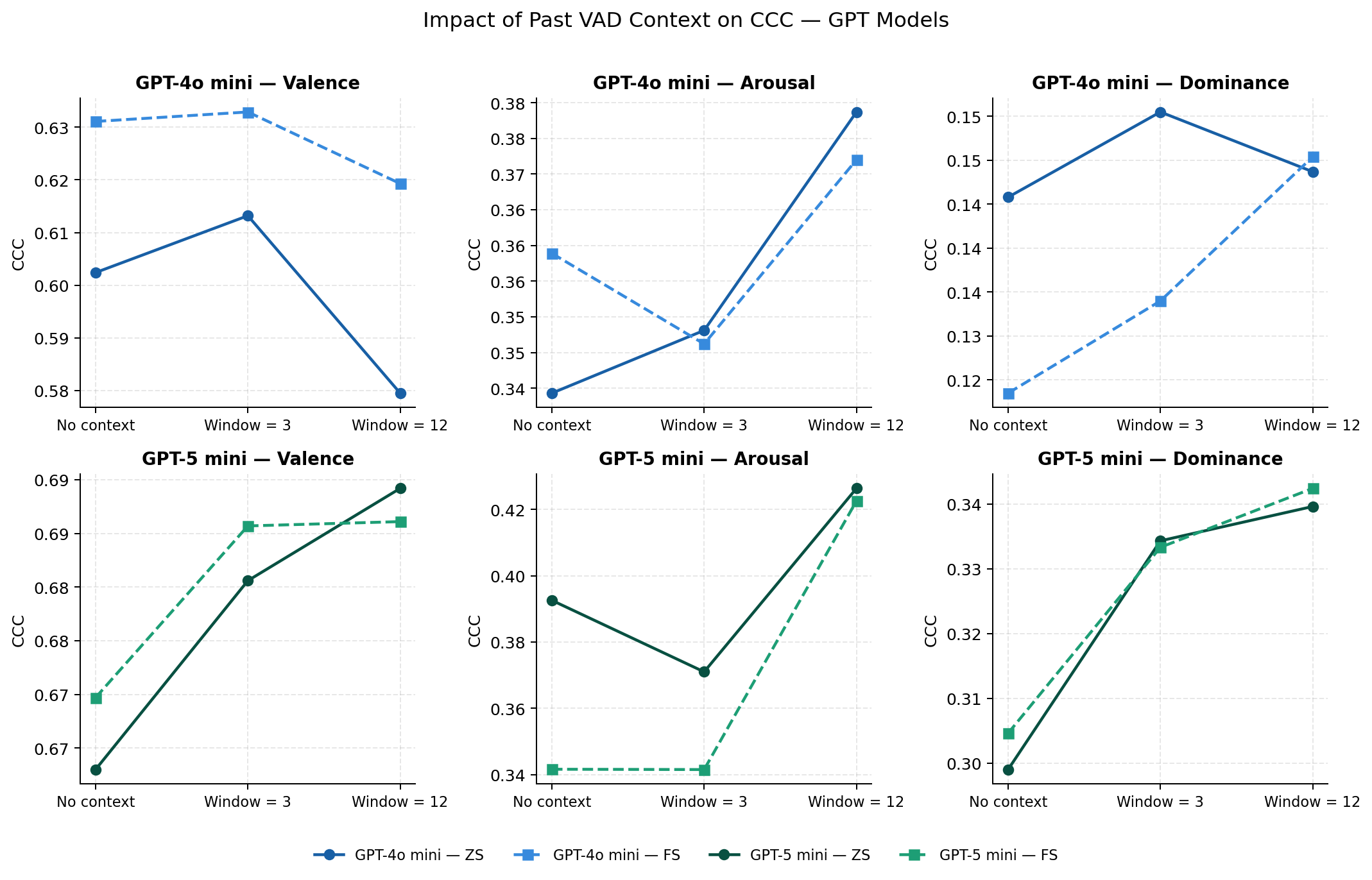}
    \caption{The Impact of Past VAD on GPT Models in Emotion Dimensional Evaluation }
    \label{fig:past_vad_gpt}
\end{figure*}

\subsubsection{LoRA Fine-tuned LLaMA Models}\label{LoRA Fine-tuned LLaMA Models}
Table \ref{tab:LLaMA_past_vad} reports the effect of past VAD context conditioning on LoRA fine-tuned LLaMA models. Unlike the GPT results, providing past VAD context generally degrades performance for fine-tuned models. For LLaMA-3.3-70B, including estimated past VAD with a 12-utterance window reduces Valence CCC from 0.7822 to 0.5368.

Importantly, this degradation does not imply that past VAD information is inherently uninformative. An oracle experiment in which ground-truth VAD scores from prior utterances are provided as context yields substantially higher performance than experiments without past VAD scores, demonstrating that temporal VAD context carries meaningful signal. The observed degradation in the estimated-VAD condition is therefore better attributed to error accumulation: when model-predicted VAD values from earlier utterances are fed back as context, any prediction errors propagate forward and corrupt the conditioning signal for subsequent utterances. 

\begin{figure*}[t]
     \centering
     \includegraphics[width =0.65\linewidth]{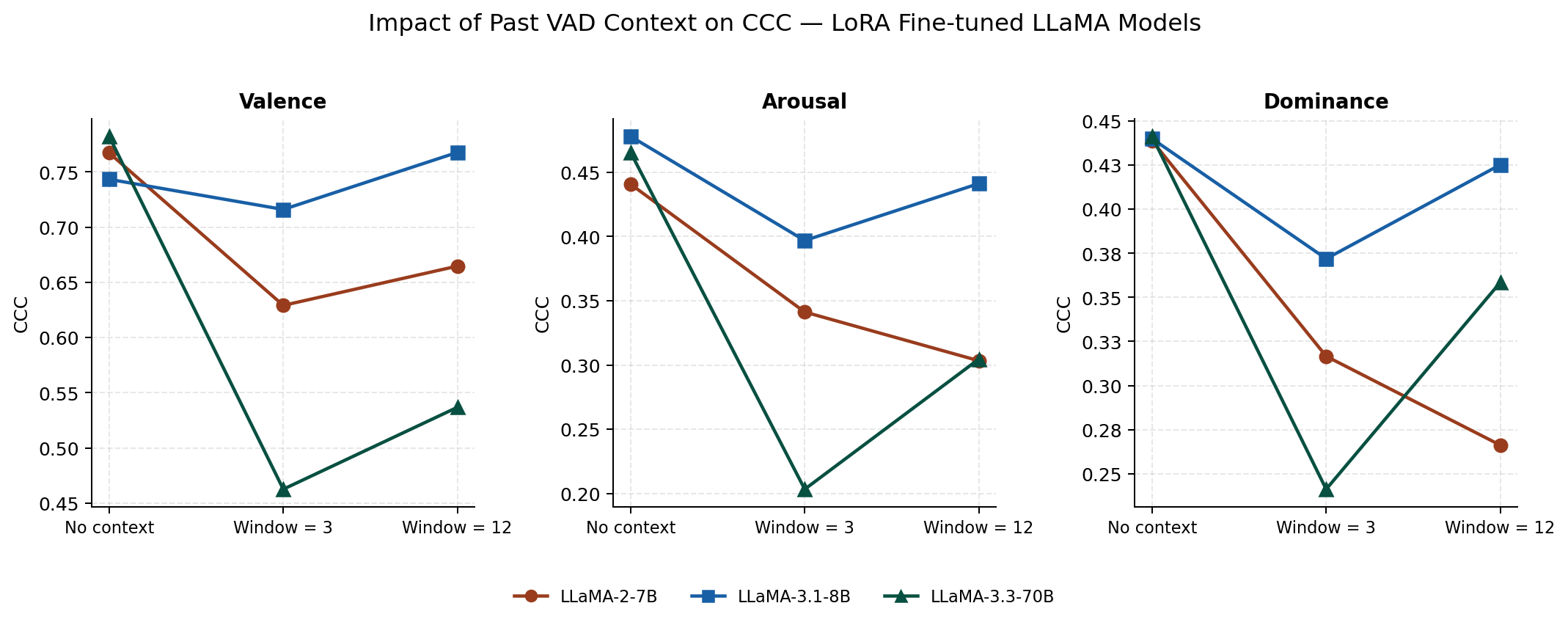}
     \caption{The Impact of Past VAD on LoRA Fine-tuned LLaMA Models in Emotion Dimensional Evaluation }
     \label{fig:past_vad_llama}
 \end{figure*}

\section{Conclusion}\label{Conclusion}
This thesis investigated the application of Large Language Models to two independent tasks in multimodal dialogue: discrete emotion recognition (ER) and dimensional emotion evaluation along the VAD continuum. Both tasks were applied to IEMOCAP using a shared input construction strategy, which combines conversation history, target utterance, and textual audio feature descriptions following the SpeechCueLLM approach.

Our results yield four principal findings. First, LoRA fine-tuning dramatically outperforms prompt engineering for both tasks. Fine-tuned LLaMA models achieve weighted F1 scores of 71.8–73.2 for discrete ER, compared to 54.7–59.6 for the best GPT prompting conditions. For VAD evaluation, fine-tuned LLaMA-3.3-70B achieves a Valence CCC of 0.7822, establishing a new state-of-the-art on the IEMOCAP dataset and outperforming the best prompt-engineered GPT result by approximately 0.11. These gains indicate that domain adaptation through parameter-efficient fine-tuning is more impactful than general instruction-following capacity for specialized emotion tasks.

Second, audio feature descriptions contribute meaningfully to performance, especially for smaller models. In the task of discrete ERC, removing audio descriptions from LoRA fine-tuned models reduces weighted F1 by 3.5–3.6 for LLaMA-2-7B and LLaMA-3.1-8B, with negligible effect on LLaMA-3.3-70B. This suggests that larger models can partially recover prosodic context from linguistic patterns, but audio information remains a cost-effective signal for smaller architectures.

Third, performance asymmetry across VAD is explained by annotator agreement. The high CCC score on Valence along the much lower scores for Arousal and Dominance mirrors their inter-annotator agreement scores hierarchy, directly linking annotation noise to model performance ceilings. 

Fourth, providing explicit past VAD values estimated by models as context helps prompt-engineered models marginally but degrades performance for LoRA fine-tuned models. This degradation is attributed to error accumulation instead of the incorporation of past VAD scores itself.

\section{Limitation}\label{Limitation}
Limitations of this work include the restriction to a single dataset (IEMOCAP), which lacks solid reliability due to its low inter-annotator agreement on the annotation of discrete emotion labels, Arousal, and Dominance. While IEMOCAP was the only available dataset satisfying all requirements after reviewing 19 candidates, the noisy data set a ceiling on the models' performance; future work should prioritize the development of additional multimodal dialogue datasets with continuous dimensional annotations. Moreover, rather than directly using a multimodal large model, we turned the raw audio recordings into textual audio feature descriptions and fed them into LLMs. Although features like volume, pitch, and speed are kept, some acoustic information is inevitably lost in this process. This might be another underlying reason for models gaining much better results in Valence than in Arousal and Dominance, for the latter two might be more acoustic information dependent.

In the future, we intend to use multimodal large models to directly process the raw audio recordings so that the acoustic information can be preserved more completely. Moreover, realizing the shortage of available data sources, we aim to establish more reliable multimodal dialogue emotion datasets. Furthermore, we hope to extend this study to text-to-speech generation with emotional control.

\section*{Acknowledgments}
We thank SAIL Lab at USC for giving us permission to IEMOCAP dataset.

\FloatBarrier
\bibliography{custom}

\FloatBarrier
\appendix

\label{appendix}
\section{Methodology Pipeline Algorithm, Details, and Examples}\label{Methodology Pipeline Details and Examples}
\begin{figure}
    \centering
    \includegraphics[width =1\linewidth]{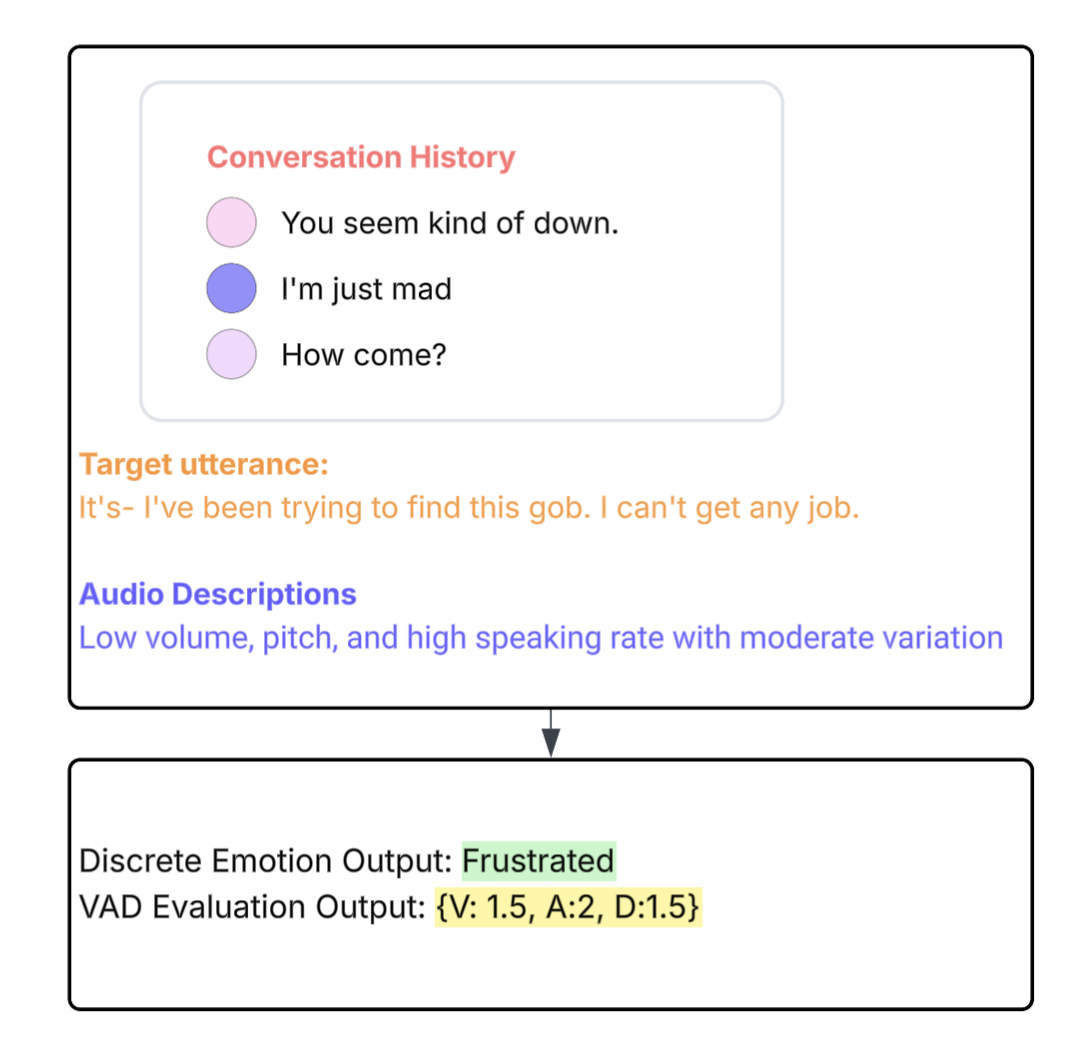}
    \caption{Example of The Proposed Pipeline}
    \label{fig:eg_pipeline}
\end{figure}

\begin{figure*}
    \centering
    \includegraphics[width =0.7\linewidth]{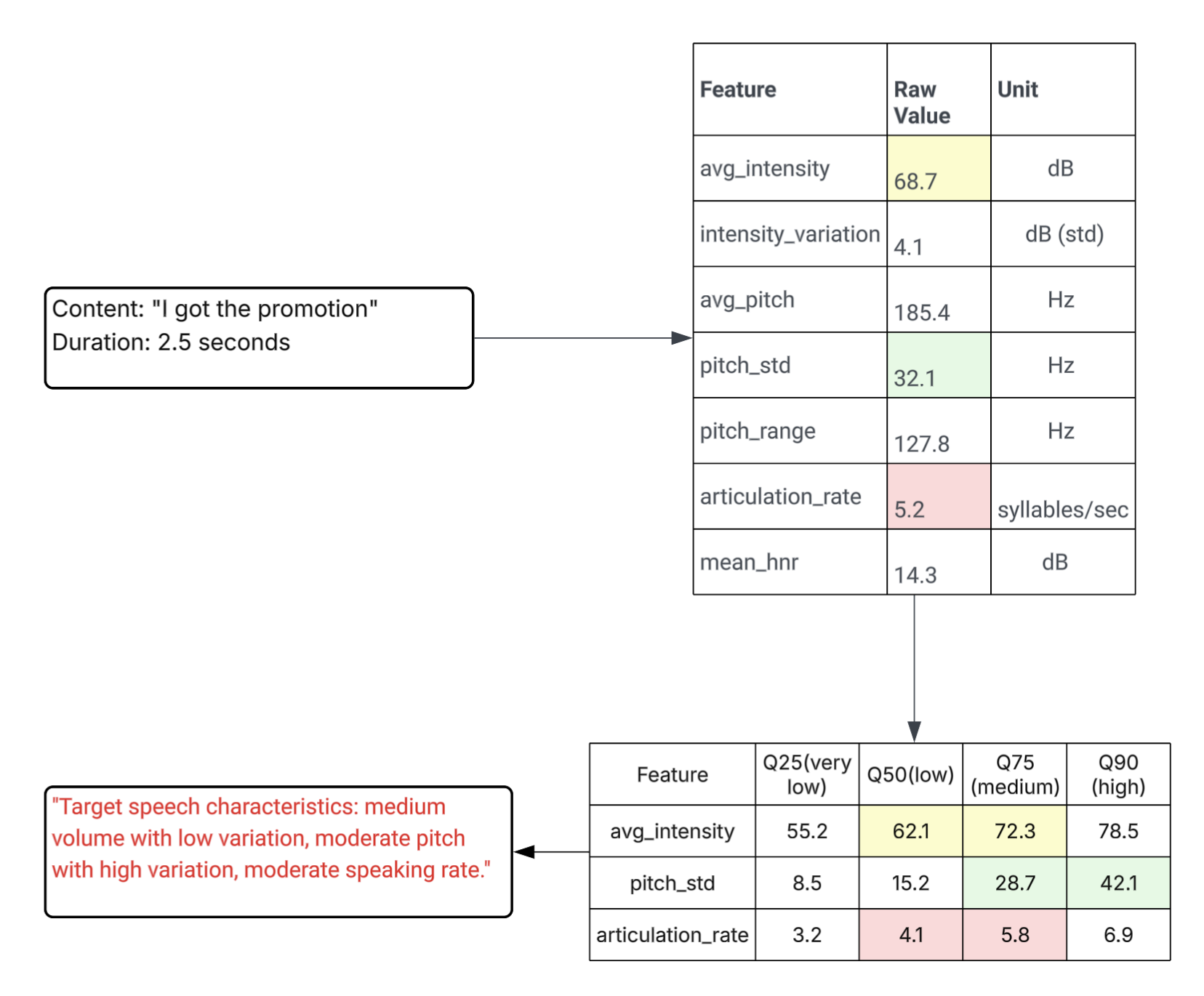}
    \caption{Example of the Audio Feature Description Generation}
    \label{fig:audio_feature_desc_eg}
\end{figure*}

Figure \ref{fig:eg_pipeline} gives an example of how the proposed pipeline will work with emotion data, and the Algorithm \ref{alg:framework} includes more details of the pipeline. Furthermore, Fig \ref{fig:audio_feature_desc_eg} visualizes the entire process of converting audio information from IEMOCAP dataset to textual audio feature description.
\begin{algorithm*}[t]
\caption{Multimodal LLM-based Emotion Evaluation}
\label{alg:framework}

\KwIn{Target utterance $u_t$, audio recording $a_t$, conversation history $\mathcal{H} = \{(s_i, u_i)\}_{i =t-W}^{t-1}$} 

\KwOut{Emotion label $\hat{y} \in \mathcal{E}$ \textbf{or} VAD scores $(\hat{v}, \hat{a}, \hat{d}) \in [1,5]^3$}

\BlankLine
\tcp{Step 1: Audio Feature Extraction}
\For{each feature $f \in \{\text{volume}, \text{pitch}, \text{speaking rate}\}$}{
    $\mu_f, \sigma_f \leftarrow \textsc{ExtractStats}(a_t, f)$\;
    $\ell_f \leftarrow \textsc{Quantize}(\mu_f,\ \{q_{0.25},\ q_{0.50},\ q_{0.75}\})$\;
    $\ell_f^{\sigma} \leftarrow \textsc{Quantize}(\sigma_f,\ \{q_{0.25},\ q_{0.50},\ q_{0.75}\})$\;
}
$d_{\text{audio}} \leftarrow \textsc{ToText}(\{\ell_f, \ell_f^{\sigma}\}_{f})$
\tcp*{e.g., ``moderate volume with high variation''}

\BlankLine
\tcp{Step 2: Prompt Construction}
$c_{\text{history}} \leftarrow \textsc{Format}(\mathcal{H})$
\tcp*{prefix each turn with speaker ID}
$p \leftarrow \textsc{AssemblePrompt}(c_{\text{history}},\ u_t,\ d_{\text{audio}},\ \tau)$
\tcp*{$\tau$: task-specific template}

\BlankLine
\tcp{Step 3: Independent Task Inference}
\eIf{task $ =$ \textsc{DiscreteER}}{
    $\hat{y} \leftarrow \text{LLM}_{\text{ER}}(p)$,\quad $\hat{y} \in \mathcal{E} = \{\text{happy, sad, neutral, angry, excited, frustrated}\}$\;
}{
    $(\hat{v},\ \hat{a},\ \hat{d}) \leftarrow \text{LLM}_{\text{VAD}}(p)$,\quad $\hat{v}, \hat{a}, \hat{d} \in \{1, 2, 3, 4, 5\}$\;
}

\BlankLine
\tcp{Step 4: (LoRA only) Model is separately fine-tuned per task}
\If{mode $ =$ \textsc{LoRA}}{
    $\text{LLM}_{\text{ER}} \leftarrow \textsc{LoRAFineTune}(\text{LLM}_{\text{base}},\ \mathcal{D}_{\text{ER}},\ r{ =}16,\ \alpha{ =}16)$\;
    $\text{LLM}_{\text{VAD}} \leftarrow \textsc{LoRAFineTune}(\text{LLM}_{\text{base}},\ \mathcal{D}_{\text{VAD}},\ r{ =}16,\ \alpha{ =}16)$\;
}

\end{algorithm*}

\section{Multimodal Emotion Dataset}\label{Multimodal Emotion Dataset}
Current multimodal emotion datasets vary significantly in their modality coverage, text types and annotation content for intensity-focused research. 
\begin{table*}[htbp]
\centering
\caption{Existing Emotion Datasets}
\label{tab:emotion_datasets}
\resizebox{\textwidth}{!}{%
\begin{tabular}{lccccc}
\toprule
\textbf{Dataset} & \textbf{Modalities} & \textbf{\# Emotions} & \textbf{Continuous Measurements} & \textbf{Dimensional Metric} & \textbf{Dialogue} \\
\midrule
IEMOCAP & T, A, V & 9 & \checkmark & VAD & \checkmark \\
MELD \cite{poria_meld_2019}& T, A, V & 7 & \texttimes & - & \checkmark \\
EmotionLines \cite{chen_emotionlines_2018} & T & 6 & \texttimes & - & \checkmark \\
DailyDialog \cite{li_dailydialog_2017} & T & 7 & \texttimes & - & \checkmark \\
CMU-MOSEI \cite{bagher_zadeh_multimodal_2018} & T, A, V & 6 & \checkmark & 0-3 Likert scale & \texttimes \\
MEISD \cite{firdaus_meisd_2020}& T, A, V & 8 & \checkmark & 1-3 Likert scale & \checkmark \\
MSP-IMPROV \cite{busso_msp-improv_2017} & T, A, V & 4 & \texttimes & - & \checkmark \\
RAVDESS \cite{singh_speech_2023}& T, A, V & 6 & \checkmark & 2 levels & \texttimes \\
CREMA-D \cite{cao_crema-d_2014} & T, A, V & 6 & \checkmark & 3 levels & \texttimes \\
MEAD \cite{wang_mead_2020} & A, V & 7 & \checkmark & 3 levels & \checkmark \\
Aff-Wild2 \cite{kollias_aff-wild2_2019}& A, V & 7 & \checkmark & VAD & \texttimes \\
GoEmotions \cite{demszky_goemotions_2020}& T & 28 & \texttimes & - & \texttimes \\
Empathetic Dialogues \cite{rashkin_towards_2019}& T & 32 & \texttimes & - & \checkmark \\
DeepDialogue \cite{koudounas_deepdialogue_2025}& T & 20 & \texttimes & - & \checkmark \\
EmotionTalk \cite{sun_emotiontalk_2025}& T, A, V & 7 & \texttimes & - & \checkmark \\
ESD \cite{zhou_emotional_2022}& T, A & 5 & \texttimes & - & \texttimes \\
Expresso \cite{nguyen_expresso_2023} & T, A & 19 & \texttimes & - & \checkmark \\
AvaMERG \cite{zhang_towards_2025-1}& T, A, V & 7 & \texttimes & - & \texttimes \\
Counselling Transcripts \cite{noauthor_cuempathy_nodate}  & T & 11 & \texttimes & - & \checkmark \\
\bottomrule
\end{tabular}%
}
T: Text, A: Audio, V: Vision; VAD: Valence-Arousal-Dominance
\end{table*}
\subsection{Dataset Coverage by Modality}
Text-only datasets like GoEmotions (58k Reddit comments, 28 emotions) and DailyDialog (13k dialogues) provide broad emotional coverage but lack multimodal context. Pure audio-visual datasets such as MEAD (40 hours) and RAVDESS (7,462 clips) include emotion intensity levels but consist primarily of acted performances. Comprehensive multimodal datasets like IEMOCAP (12 hours audiovisual data) and CMU-MOSEI (23.5k utterances) combine text, audio, and visual modalities with both categorical and dimensional annotations.

\subsection{Emotion Continuous Dimensional Annotations}
Few datasets explicitly model emotion on continuous dimensions, which is crucial for quantitative evaluation. IEMOCAP provides VAD system ratings (valence, arousal, dominance), while MEAD categorizes intensity as neutral, weak, medium, and strong. MEISD offers explicit intensity levels (1-3) across eight emotion categories. However, most datasets including MELD, EmotionLines, and GoEmotions provide only categorical emotion labels without dimensional emotion measures.

\subsection{Conversational Versus Non-Conversational Context}
Datasets vary in their contextual structure. Conversational datasets like MELD, IEMOCAP, and MEISD preserve turn-taking dynamics and contextual emotion flow. Non-conversational datasets such as CMU-MOSEI (YouTube monologues) and actor-based datasets (MEAD, RAVDESS, CREMA-D) focus on isolated emotional expressions without interactive context.\\

The predominance of categorical over continuous dimensional annotations limits research in emotion quantitative evaluation. For multimodal emotion dimensional evaluation research, IEMOCAP and MEISD represent the most suitable resources. However, due to the unavailability of MEISD dataset, our final and only available choice is IEMOCAP dataset.
\begin{table*}[h]
\centering
\caption{Summary Statistics of the IEMOCAP Dataset}
\label{tab:iemocap_stats}
\begin{tabular}{|l|l|}
\hline
\textbf{Property} & \textbf{Value} \\\hline

Number of sessions          & 5 dyadic sessions \\\hline
Total dialogues             & 151 \\\hline
Total utterances            & 10,086 \\\hline
Avg. utterances per dialogue & $\approx$ 66 \\\hline
Total audio duration        & $\approx$ 12 hours \\\hline
Available modalities        & Audio, Video, Text, Motion Capture \\\hline
Discrete emotion categories used & 6 (happy, sad, neutral, angry, excited, frustrated) \\\hline
Utterances after filtering  & 7,433 \\
\bottomrule
\end{tabular}
\end{table*}
 \begin{figure}
     \centering
     \includegraphics[width =0.75\linewidth]{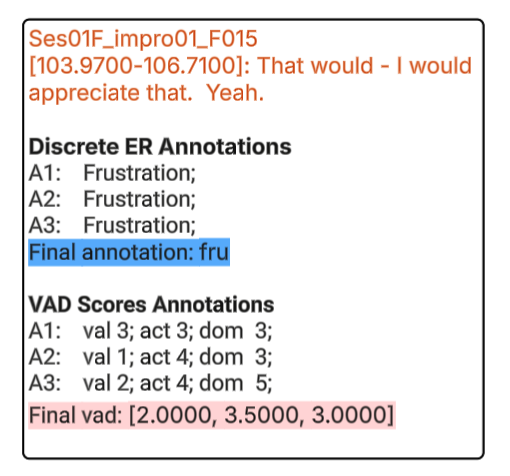}
     \caption{Example of IEMOCAP Dataset}
     \label{fig:data_eg}
 \end{figure}

\section{Prompt Templates}\label{Prompt Templates}
Reasoning was required for prompts for both tasks to boost the performance.
\subsection{Discrete ER Prompt}\label{Discrete ER Prompt}
The zero-shot prompt for discrete ER follows this structure:

\begin{center}
\begin{quote}
\ttfamily
[System] You are an expert in emotional analysis for dialogues. Select one emotion label from \{happy, sad, neutral, angry, excited, frustrated\} and respond in strict JSON format.

[User] Conversation history: \{\{conversation\_history\}\} \\
Target utterance: \{\{target\_utterance\}\} \\
Audio features: \{\{audio\_features\}\}

Select the single emotion that best represents the dominant emotion of the target utterance. Respond in JSON: \\
\{"emotion\_label": "...", "reasoning": "..."\}
\end{quote}
\end{center}

Few-shot prompts prepend a fixed set of annotated examples (with input–output pairs) before the target utterance. The examples are selected to cover all six emotion categories and to represent cases where audio features are diagnostic.

\subsection{VAD Evaluation Prompt}\label{VAD Evaluation Prompt}
The zero-shot prompt for VAD evaluation follows the same structure, with the output format modified to return three integer scores:

\begin{center}
\begin{quote}
\ttfamily
[System] You are an expert in dimensional emotion analysis. Rate the target utterance on Valence (V), Arousal (A), and Dominance (D), each on a scale of 1 to 5, where higher values indicate more positive, more activated, and more dominant respectively. Respond in strict JSON format.

[User] Conversation history: \{\{conversation\_history\}\} \\
Target utterance: \{\{target\_utterance\}\} \\
Audio features: \{\{audio\_features\}\}

Respond in JSON: \\
\{"valence": <1-5>, "arousal": <1-5>, "dominance": <1-5>, "reasoning": "..."\}
\end{quote}
\end{center}
\subsection{Detailed Results for the Impact of Past VAD Values on VAD Evaluation Performance}\label{VAD Evaluation Prompt}
Table \ref{tab:gpt_past_vad12} and Table \ref{tab:LLaMA_past_vad} provide detailed results of the ablation study for the impact of past VAD values on VAD evaluation performance.
 \begin{table*}[ht]
\centering
\caption{VAD Evaluation Performance Comparison Across OpenAI Models and Prompting Strategies with Past VAD Values }
\label{tab:gpt_past_vad12}
\scalebox{0.9}
{\begin{tabular}{l|cc|cc|cc|cc}
\toprule
&  \multicolumn{4}{c}{Context Window = 12} & \multicolumn{4}{c}{Context Window = 3} \\
\cmidrule(lr){2-5}\cmidrule(lr){6-9}
&  \multicolumn{2}{c}{GPT-4o mini} & \multicolumn{2}{c}{GPT-5 mini}&  \multicolumn{2}{c}{GPT-4o mini} & \multicolumn{2}{c}{GPT-5 mini} \\
\cmidrule(lr){2-3} \cmidrule(lr){4-5}\cmidrule(lr){6-7}\cmidrule(lr){8-9}
Dimension  & Zero-shot & Few-shot & Zero-shot & Few-shot& Zero-shot & Few-shot & Zero-shot & Few-shot \\
\midrule

Valence       & 0.5795 & 0.6193 & \textbf{0.6892} & 0.6861& 0.6132 & 0.6329 & 0.6806 & 0.6857 \\

Arousal       & 0.3787 & 0.3720 & \textbf{0.3959} & 0.3460& 0.3481 & 0.3462 & 0.3710 & 0.3415 \\

Dominance     & 0.1487 & 0.1504 & 0.3396 & \textbf{0.3424}& 0.1555 & 0.1340 &0.3343 & 0.3333 \\

Overall  & 0.3690 & 0.3806 & \textbf{0.4749} & 0.4582& 0.3723 & 0.3710 & 0.4620 & 0.4535 \\
\bottomrule
\end{tabular}}
\end{table*}
 \begin{table*}[ht]
\centering
\caption{VAD Evaluation Performance Comparison Across LLaMA Models with LoRA Finetuning With Past VAD}
\label{tab:LLaMA_past_vad}
\scalebox{0.9}{\begin{tabular}{l|ccc|ccc}
\toprule
&\multicolumn{3}{c}{Context Window = 12}&\multicolumn{3}{c}{Context Window = 3} \\
\cmidrule(lr){2-4} \cmidrule(lr){5-7}
 Dimension&2-7B&3.1-8B&3.3-70B&2-7B&3.1-8B&3.3-70B\\
\midrule

Valence       &0.6647&\textbf{0.7677}&0.5368&0.6291&0.7160 & 0.4626 \\

Arousal       &0.3033&\textbf{0.4411}&0.3046&0.3412&0.3968&0.2034 \\

Dominance       &0.2662&\textbf{0.4252}&0.3585&0.3166&0.3719& 0.2412\\

Overall       &0.4114&\textbf{0.5447}&0.4000&0.5489&0.4949 &0.3024\\
\bottomrule
\end{tabular}}
\end{table*}

\end{document}